\documentclass[10pt,twocolumn,letterpaper]{article}

\usepackage{iccv}
\usepackage{times}
\usepackage{graphicx}
\usepackage{subfig}
\graphicspath{{./figures/}}
\usepackage{amsmath}
\usepackage{amssymb}
\usepackage{leftidx}
\usepackage{multirow}
\usepackage{makecell}
\usepackage{pgfplots}
\pgfplotsset{compat=1.5}
\usepackage{tikz}
\usepackage{bm}
\usepackage{algorithm}
\usepackage[noend]{algpseudocode}
\usepackage{bbold}
\usepackage{xurl}

\newcommand\todo[1]{\textcolor{red}{#1}}
\usepackage{booktabs}

\iccvfinalcopy 

\def\iccvPaperID{6737} 

\ificcvfinal\fi
\begin{document}

\title{Looking Fast and Slow: Memory-Guided Mobile Video Object Detection}

\author{Mason Liu\\
Cornell University\thanks{This work was done while interning at Google.} \\
\texttt{\small masonliuw{@}cs.cornell.edu}\\
\and Menglong Zhu \enskip Marie White \enskip Yinxiao Li \enskip Dmitry Kalenichenko \\
Google AI\\
\texttt{\small \{menglong,mariewhite,yinxiao,dkalenichenko\}{@}google.com}\\
}

\maketitle
\thispagestyle{empty}

\begin{abstract}
With a single eye fixation lasting a fraction of a second, the human visual system is capable of forming a rich representation of a complex environment, reaching a holistic understanding which facilitates object recognition and detection. This phenomenon is known as recognizing the "gist" of the scene and is accomplished by relying on relevant prior knowledge. This paper addresses the analogous question of whether using memory in computer vision systems can not only improve the accuracy of object detection in video streams, but also reduce the computation time. By interleaving conventional feature extractors with extremely lightweight ones which only need to recognize the gist of the scene, we show that minimal computation is required to produce accurate detections when temporal memory is present. In addition, we show that the memory contains enough information for deploying reinforcement learning algorithms to learn an adaptive inference policy. Our model achieves state-of-the-art performance among mobile methods on the Imagenet VID 2015 dataset, while running at speeds of up to 70+ FPS on a Pixel 3 phone.\footnote{Source code will be released under \cite{mobilelstd}}
\end{abstract}

\section{Introduction}
Recent advances in image object detection have followed a trend of increasingly elaborate convolutional neural network \cite{Krizhevsky, Simonyan1, Szegedy1, he2016deep} designs to improve either accuracy or speed. Though accuracy was initially the primary concern and continues to be a key metric \cite{Girshick, Hek3, RenS, liu2016ssd, DaiJ, lin2017feature, lin2017focal}, the importance of improving the speed of these models has steadily risen as deep learning techniques have been increasingly deployed in practical applications. On the far end of the speed spectrum, substantial work \cite{Howard, sandler2018mobilenetv2, ZhangX, ma2018shufflenet, WuJ, jacob2018quantization, zhu2018towardsmobile} has been done on allowing neural networks to run on mobile devices, which represent an environment with extreme computation and energy constraints. Despite significant advances, the ultimate goal of being able to run neural networks in real-time on mobile devices without substantial accuracy loss has yet to be achieved by any single-frame detection model.

\begin{figure}[t!]
\begin{center}
   \includegraphics[width=.95\linewidth]{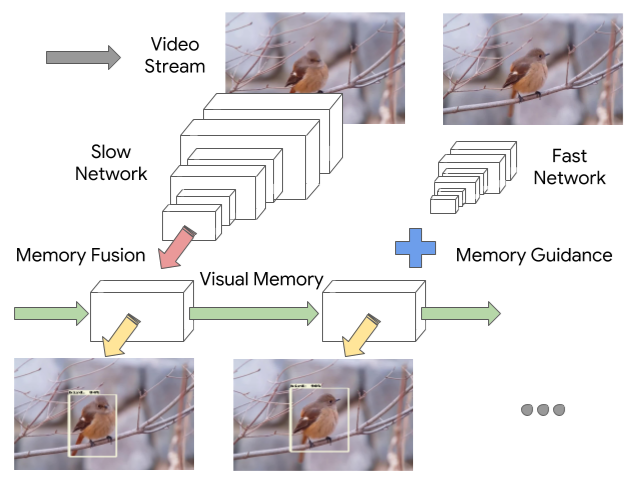}
\end{center}
\vspace{-1em}
   \caption{An illustration of our proposed memory-guided interleaved model. Given a video stream, a visual memory module fuses very different visual features produced by fast and slow feature extractors across frames to generate detections in an online fashion.}
\label{fig:key_fig}
\end{figure}

However, the human visual system provides intuition that such a result should be achievable, as experiments have shown that humans can process basic information about a scene at a single glance, which notably includes recognizing a few categories of objects \cite{oliva2005gist}. One critical advantage of human vision is that it does not operate on single images, but rather a stream of images. By introducing a temporal dimension to the problem, humans can rely on contextual cues and memory to supplement their understanding of the image. This work examines the question of whether neural networks are similarly able to perform video object detection with very little computation when assisted by memory.

A key observation is that since adjacent video frames tend to be similar, running a single feature extractor on multiple frames is likely to result in mostly redundant computation. Therefore, a simple idea is to keep a memory of previously computed features and only extract a small amount of necessary features from new frames. This parallels the role of gist in the human visual system in that both require minimal computation and rely on memory to be effective. To follow this idea, a system requires multiple feature extractors-- an accurate extractor initializes and maintains the memory, while the other rapidly extracts features representing the gist of new images.

From these observations, we propose a simple yet effective pipeline for efficient video object detection, illustrated in Figure \ref{fig:key_fig}. Specifically, we introduce a novel interleaved framework where two feature extractors with drastically different speeds and recognition capacities are run on different frames. The features from these extractors are used to maintain a common visual memory of the scene in the form of a convolutional LSTM (ConvLSTM) layer, and detections are generated by fusing context from previous frames with the gist from the current frame. Furthermore, we show that the combination of memory and gist contains within itself the information necessary to decide when the memory must be updated. We learn an \textit{interleaving policy} of when to run each feature extractor by formulating the task as a reinforcement learning problem.

While prior works, notably flow-based methods \cite{ZhuX1, zhu2018towardsmobile}, also provide approaches for fast video object detection based on interleaving fast and slow networks, these approaches are based on the CNN-specific observation that intermediate features can be warped by optical flow. Meanwhile, our method relies on the biological intuition that fast, memory-guided feature extractors exist in the human visual system. This intuition translates naturally into a simple framework which is not dependent on optical flow. Our method runs at an unprecedented 72.3 FPS post-optimization on a Pixel 3 phone, while matching state-of-the-art mobile performance on the Imagenet VID 2015 benchmark.

In summary, this paper's contributions are as follows:
\begin{itemize}
  \itemsep0em
  \item We present a memory-guided interleaving framework where multiple feature extractors are run on different frames to reduce redundant computation, and their outputs are fused using a common memory module.
  \item We introduce an adaptive interleaving policy where the order to execute the feature extractors is learned using Q-learning, which leads to a superior speed/accuracy trade-off.
  \item We demonstrate on-device the fastest mobile video detection model known to date at high accuracy level.
\end{itemize}

\section{Related Work}
Video object detection has received much attention in recent years. Current methods focus on extending single-image detection approaches by leveraging the temporal properties of videos to improve accuracy and speed. These approaches can be roughly categorized into three families.

\subsection{Postprocessing Methods}
Initial work for extending single-image detection to the video domain usually centered on a postprocessing step where per-frame detections are linked together to form tracks, and detection confidences are modified based on other detections in the track. Seq-nms \cite{HanW} finds tracks via dynamic programming and boosts the confidence of weaker predictions. TCNN \cite{KangK1, KangK2} provides a pipeline with optical flow to propagate detections across frames and a tracking algorithm to find tubelets for rescoring. These early approaches yielded sizeable performance improvements, but did not fundamentally change the underlying per-frame detection process, which limited their effectiveness.

\subsection{Feature Flow Methods}
Later, Zhu et al. \cite{ZhuX1} discovered that intermediate features in a convolutional neural network could be directly propagated between video frames via optical flow. The DFF framework \cite{ZhuX1} demonstrated that it is sufficient to compute detections on sparse keyframes and perform feature propagation on all other frames by computing optical flow, which is substantially cheaper. FGFA \cite{ZhuX2} showed that this idea can also be used to improve accuracy if per-frame detections are densely computed and features from neighboring frames are warped to the current frame and aggregated with weighted averaging. Impression networks \cite{hetang2017impression} balance speed and accuracy by using sparse keyframes but retaining an "impression feature" which is aggregated across keyframes and stores long-term temporal information. Further work by Zhu et al. \cite{zhu2018towards} introduces efficient feature aggregation as well as a measure of feature quality after warping, which is used to improve keyframe selection and sparsely replace poorly warped features. 

This paradigm has also been applied to mobile-focused video object detection, which is particularly relevant to this paper. In \cite{zhu2018towardsmobile}, flow-guided feature propagation is used on a GRU module with very efficient feature extractor and flow networks to demonstrate that flow-based methods are viable in computationally constrained environments. Our work also applies to the mobile setting, but by interleaving specialized feature extractors rather than using flow to propagate features, we remove the dependence on optical flow and hence the need for optical flow training data and the additional optical flow pre-training stage.

\subsection{Multi-frame Methods}
A third class of video object detection methods involve methods which explicitly process multiple frames of the video simultaneously. D\&T \cite{feichtenhofer2017detect} combines detection and tracking by adding an RoI tracking operation and loss on pairs of frames, while STSN \cite{bertasius2018object} uses deformable convolutions to sample features from adjacent frames. Chen et al. \cite{chen2018optimizing} propose using a scale-time lattice to generate detections in a coarse-to-fine manner. Though D\&T and Chen et al.'s methods can improve detection speed by sampling sparse keyframes and propagating results to intermediate frames, all of these works are still focused on high-accuracy detection and are nontrivial to generalize to a mobile setting. Our approach also extracts features from each frame rather than entirely propagating results from keyframes, which allows access to a greater quantity of information. 

\subsection{Adaptive Keyframe Selection}
There have been a variety of methods to select keyframes when sparsely processing videos. These methods range from fixed intervals \cite{ZhuX1, zhu2018towardsmobile} to heuristics \cite{zhu2018towards, li2018low} to learned policies \cite{li2017dynamic, mahasseni2017budget, supancic2017tracking, huang2017learning, ying2017depth}. Most of these works address the problems of semantic segmentation and tracking, and adaptive keyframe selection has been less explored in video object detection. We propose a different formulation for constructing a learned adaptive policy by leveraging the information contained in our memory module, creating a complete and principled detection pipeline.

\section{Approach}

\begin{figure}[b]
\begin{center}
\includegraphics[width=\columnwidth]{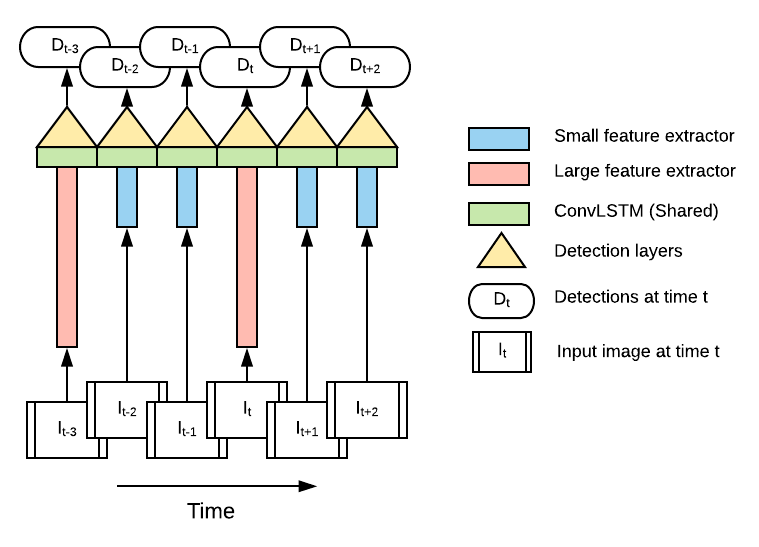}
\end{center}
\vspace{-2em}
  \caption{An illustration of the proposed interleaved model with heavy and lightweight feature extractors using a fixed interleave policy with $\tau=2$.}
\label{fig:Interleaved}
\end{figure}

\subsection{Interleaved Models}
Our paper addresses the task of video object detection. For this task, we must generate frame-level bounding boxes and class predictions on each frame of a video $\mathcal{V} = \{I_0, I_1,\ldots I_n\}$. We further restrict our task to an online setting where only  $\{I_0, I_1,\ldots I_k\}$ are available when generating detections for the $k$-th frame.

The primary contribution of this paper is an interleaved model framework where multiple feature extractors are run sequentially or concurrently. These frame-level features are then aggregated and refined using a memory mechanism. Finally, we apply SSD-style \cite{liu2016ssd} detection on the refined features to produce bounding box results.

Each of the these steps can be defined as a function. Let the $m$ feature extractors be $\mathbf{f_i}: \mathbb{R}^I \to \mathbb{R}^{F} \mid_{i=0}^m$,  mapping the image space to separate feature spaces in $\mathbb{R}^{F}$. The memory module $\mathbf{m}: \mathbb{R}^F \times \mathbb{R}^S \to \mathbb{R}^R \times \mathbb{R}^S$, maps features from $\mathbf{f}$ and an internal state representation to a common, refined feature space while also outputting an updated state. The SSD detector $\mathbf{d}: \mathbb{R}^R \to \mathbb{R}^{D}$ maps refined features to final detection anchor predictions.

The use of multiple feature extractors has several advantages. Different feature extractors can specialize on different image features, creating a temporal ensembling effect. Since we are focused on the mobile setting, we study the case where the features extractors have drastically different computational costs, which dramatically decreases the runtime of the model. In particular, the remainder of our paper focuses on the case where $m=2$, with $\mathbf{f_0}$ optimized for accuracy and $\mathbf{f_1}$ optimized for speed.

To obtain detection results $D_k$ on the $k$-th frame given the previous frame's state $s_{k-1}$, run $\mathbf{m}(\mathbf{f_i}(I_k), s_{k-1})$ to obtain a feature map $M_k$ and the updated state $s_k$. Then, $D_k = \mathbf{d}(M_k)$. This is shown in Figure \ref{fig:Interleaved}. Note that running any feature extractor (i.e. choosing any valid $i$) will yield valid detection results, but the quality of the detections and the updated state representation will vary. One crucial problem is finding an interleaving policy such that the amortized runtime of our method is similar to $\mathbf{f_1}$ while retaining the accuracy of exclusively running $\mathbf{f_0}$. A simple fixed interleaving policy involves defining a hyperparameter $\tau$, the interleave ratio, and running $\mathbf{f_0}$ after $\mathbf{f_1}$ is run $\tau$ times. Though we find that even this simple policy achieves competitive results, we also present a more advanced learned policy in section \hyperref[sec:3.4]{3.4}.

The architecture of $\mathbf{f_0}$ is a standard MobileNetV2 \cite{sandler2018mobilenetv2} with a depth multiplier of $1.4$ and an input resolution of $320 \times 320$. $\mathbf{f_1}$ also uses a MobileNetV2 architecture with a depth multiplier of $0.35$ and a reduced input resolution of $160\times 160$. We remove the striding on the last strided convolution so that the output dimensions match. The SSDLite layers are similar to \cite{sandler2018mobilenetv2}, with the difference that the SSD feature maps have a constant channel depth of 256 and share the same convolutional box predictor. We also limit the aspect ratios of the anchors to $\{1, 0.5, 2.0\}$.

\subsection{Memory Module}\label{sec:3.2}
Our method requires a memory module for aggregating features from the two extractors across timesteps, especially to augment features from the small network with previous temporal context. Though Liu and Zhu \cite{liu2018mobile} showed that LSTMs can be used to propagate temporal information for object detection, our memory module serves an additional purpose of fusing features from different feature extractors and feature spaces, posing an additional challenge. Moreover, we require this mechanism to be extremely fast, as it must be executed on all frames. To this end, we modify an LSTM cell to be faster and better at preserving long-term dependencies.

\begin{figure}[ht]
\begin{center}
  \includegraphics[width=\linewidth]{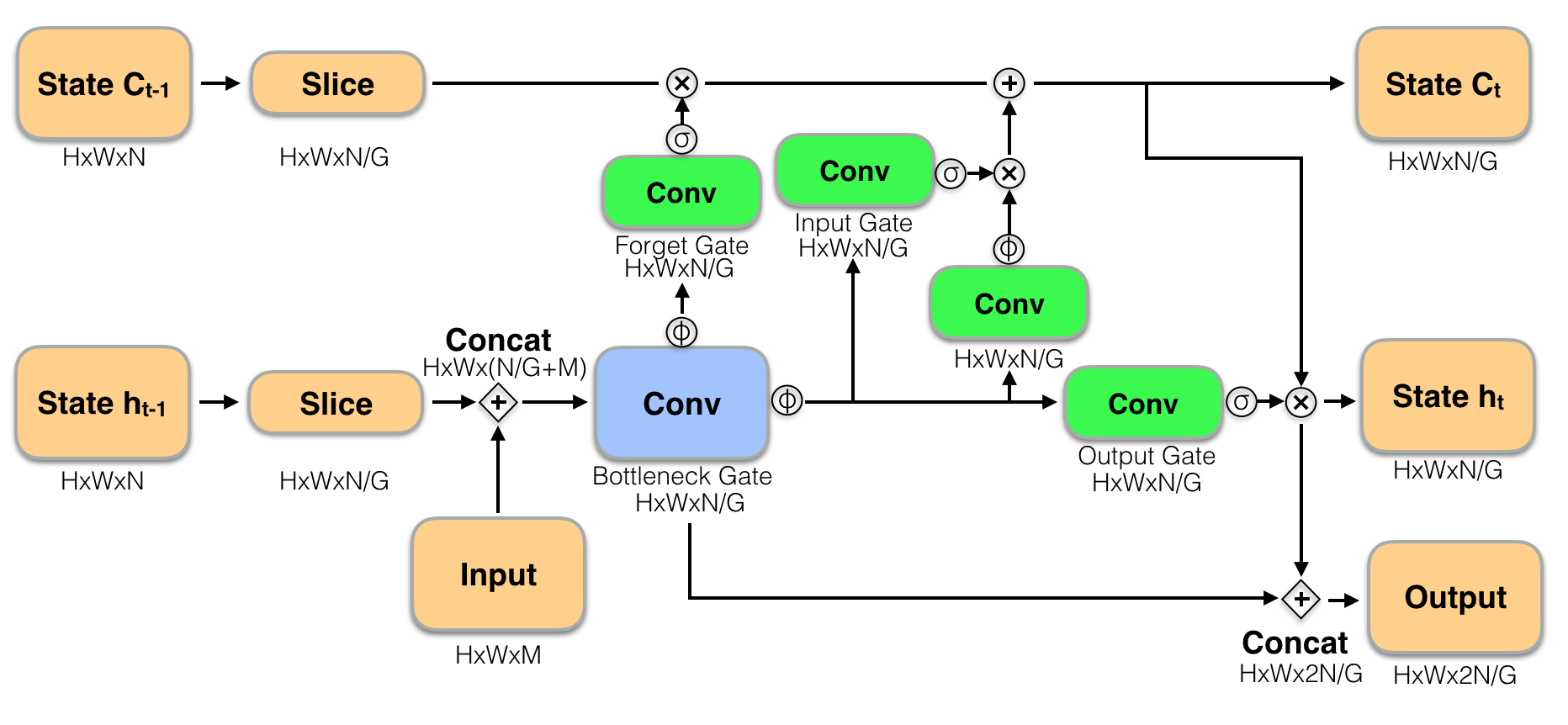}
\vspace{-2em}
\end{center}
  \caption{Detailed illustration of one group of our speed-optimized LSTM cell. The illustrated operations are performed once for each of $G$ groups.}
\label{fig:fast-lstm}
\end{figure}

To improve the speed of the standard LSTM, we make three modifications. We adopt the bottlenecking proposed in \cite{liu2018mobile} and add a skip connection between the bottleneck and output so that the bottleneck is part of the output. We also divide the LSTM state into groups and use grouped convolutions to process each one separately. Given the previous state $h_{t-1}$ and input feature map $x_t$, we partition the state channel-wise into $G$ equal partitions $^{1}h_{t-1}, ^{2}h_{t-1} \ldots ^{G}h_{t-1}$. We concatenate each partition with $x_t$ and compute bottlenecked LSTM gates $b_t, f_t, i_t, o_t$ as in \cite{liu2018mobile}. The updated LSTM states $^{g}c_t, ^{g}h_t$ are also computed the same way, but are now just slices of the final updated state. We generate the output slice with a skip connection:
\begin{align*}
&^{g}M_t = [^{g}h_t, b_t],
\end{align*}
where brackets denote concatenation. Finally, concatenate the slices channel-wise to obtain $c_t$, $h_t$, and $M_t$. This is visualized in Figure \ref{fig:fast-lstm}. The grouped convolutions provide a speed-up by sparsifying layer connections, while the skip connection allows less temporally relevant features to be included in the output without being stored in memory. This reduces the burden on the memory and allows us to scale down the state dimensions, whereas other LSTM variants suffer substantial accuracy drops if the state is too small \cite{liu2018mobile}. In our model, we choose $G=4$ and use a $320$-channel state. The speed benefits of our modified LSTM are detailed in Table \ref{tab:lstm}.

We also observe that one inherent weakness of the LSTM is its inability to completely preserve its state across updates in practice. The sigmoid activations of the input and forget gates rarely saturate completely, resulting in a slow state decay where long-term dependencies are gradually lost. When compounded over many steps, predictions using the $\mathbf{f_1}$ degrade unless $\mathbf{f_0}$ is rerun.

We propose a simple solution to this problem by simply skipping state updates when $\mathbf{f_1}$ is run, i.e. the output state from the last time $\mathbf{f_0}$ was run is always reused. This greatly improves the LSTM's ability to propagate temporal information across long sequences, resulting in minimal loss of accuracy even when $\mathbf{f_1}$ is exclusively run for tens of steps.

\begin{table}[t]
\begin{tabular}{c | c | c }
Recurrent Layer Type & mAP & MAC (M) \\ 
\toprule [0.2em]
ConvLSTM & \textbf{63.40} & 317 \\
ConvGRU & 63.20 & 240 \\
Bottleneck LSTM \cite{liu2018mobile} & 61.92 & 186 \\
Ours & \textbf{63.37} & \textbf{39} \\
\bottomrule [0.2em]
\end{tabular}
\centering
\caption{Performance of a MobilenetV2-SSDLite with a recurrent layer as in \cite{liu2018mobile}. Each RNN variant has a 1024-channel input and 640-channel output. mAP@0.5IOU is reported on a subset of Imagenet VID validation containing a random sequence of 20 frames from each video. MAC contains only multiply-adds from the RNN.}
\label{tab:lstm}
\end{table}

\subsection{Training Procedure}
Our training procedure consists of two phases. First, we pretrain our interleaved model without detection layers on Imagenet classification in order to obtain a good initialization of our LSTM weights. To adapt the network for classification, we remove the detection layers $\mathbf{d}$ and add one average pooling and fully connected layer immediately after the LSTM, followed by a softmax classifier. During training, we duplicate each frame three times and unroll the LSTM to three steps. At each step, we uniformly select a random feature extractor to run.

Next, we perform SSD detection training. Again, we unroll the LSTM to six steps and uniformly select a random feature extractor at each step. We train on a mix of video and image data. For image data, we augment the image by cropping a specific region at each step and shifting the crop between steps to mimic translations and zooms, in order to aid the model in learning the relation between motion and box displacement. Otherwise, the training procedure is similar to that of the standard SSD \cite{liu2016ssd}. We use a batch size of $12$ and a learning rate of $0.002$ with cosine decay.

\subsection{Adaptive Interleaving Policy}
\label{sec:3.4}
\begin{figure}[t]
\begin{center}
  \includegraphics[width=1.0\linewidth]{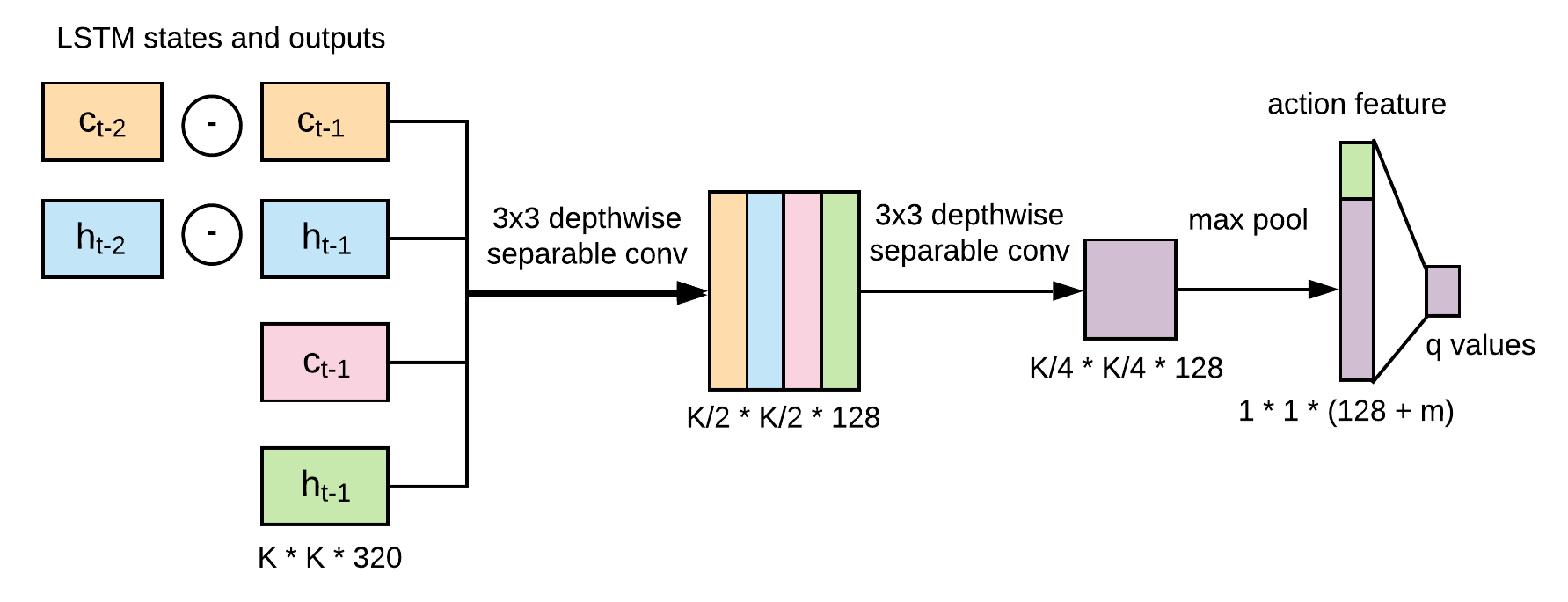}
\end{center}
\vspace{-1em}
  \caption{Our adaptive interleaved model uses an ultra-lightweight policy network to decide which feature extractor to run at each time step.}
\label{fig:rl}
\end{figure}

Though we show that a simple interleaving policy already achieves competitive results, a natural question is whether it is possible to optimize the interleaving policy to further improve results. We propose a novel approach for learning an adaptive interleaving policy using reinforcement learning. The key observation is that in order to effectively aid the smaller network, the memory module must contain some measure of detection confidence, which we can leverage as part of our interleaving policy. Therefore, we construct a policy network $\pi$ which examines the LSTM state and outputs the next feature extractor to run, as shown in Figure \ref{fig:rl}. Then, we train the policy network using Double Q-learning (DDQN) \cite{van2016deep}.

To formulate a reinforcement learning problem, it is necessary to define an action space, a state space, and a reward function. The action space consists of $m$ actions, where action $a$ corresponds to running $\mathbf{f_a}$ at the next timestep. We denote the state as:
\begin{align}
    S = (c_t, h_t, c_t - c_{t-1}, h_t - h_{t-1}, \eta_t)
\end{align}
which includes the current LSTM states $c_t$ and $h_t$, as well as their changes during the current step $(c_t - c_{t-1})$ and $(h_t - h_{t-1})$. We also add an action history term $\eta$, so the policy network is aware of its previous actions and can avoid running $\mathbf{f_0}$ excessively. The action history is a binary vector of length 20. For all $k$, the $k$-th entry of $\eta$ is $1$ if $\mathbf{f_1}$ was run $k$ steps ago and $0$ otherwise. \todo{}

Our reward function must reflect our aim to find a balance between running $\mathbf{f_1}$ as frequently as possible while maintaining accuracy. Therefore, we define the reward as the sum of a speed reward and an accuracy reward. For the speed reward, we simply define a positive constant $\gamma$ and give $\gamma$ reward when $\mathbf{f_1}$ is run. For the accuracy reward, we compute the detection losses after running each feature extractor and take the loss difference between the minimum-loss feature extractor and the selected feature extractor. The final reward can be expressed as: \\
\begin{align}
R(a) = \begin{cases} \label{eq:reward}
\min\limits_i L(D^i) - L(D^0) & a = 0 \\
\gamma + \min\limits_i L(D^i) - L(D^1) & a = 1 \,, \\
\end{cases}
\end{align}
where $L(D^i)$ denotes the loss for detections $D^i$ using features from $\mathbf{f_i}$. 

Our policy network is a lightweight convolutional neural network which predicts the Q-value of each state-action pair given the state. We first perform a grouped convolution using each of the four feature maps in $S$ as separate groups. Then, we perform a depthwise separable convolution and use max pooling to remove the spatial dimensions. Finally, we concatenate the action feature vector and apply a fully connected layer to obtain $m$ outputs, the Q-values for each state-action pair. The architecture is illustrated in Figure \ref{fig:rl}.

To train the policy network, we generate batches of $(S_t, a, S_{t+1}, R_t)$ examples by running the interleaved network in inference mode. Though the entire system could potentially be trained end-to-end, we simplify the training process by using pretrained weights for the base interleaved model and freezing all weights outside of the policy network. After obtaining the batched examples, we use standard DDQN with experience replay as described in \cite{van2016deep}. The training process is detailed in Algorithm \ref{alg:training}.

\begin{algorithm}
\caption{Adaptive Interleaving Training}
\renewcommand{\algorithmicrequire}{\textbf{Define:}}
\begin{algorithmic}[1]
\Require action $a$, timestep $t$, action history $\eta$, LSTM state $s$, observation $S$, reward $R$, policy $\pi$
\Repeat
\label{alg:training}
\State \text{sample video frames} $I_1, \ldots I_k$
\State $a_0 \gets 0; t \gets 0; \eta_0 \gets \mathbf{0}$
\While{$t < k$}
\For{every feature extractor $\mathbf{f_i}$}
\State $M^i_t, s^i_t \gets \mathbf{m}(\mathbf{f_i}(I_t), s_{t-1})$
\If{$i = a_t$}
\State $s_t \gets s^i_t$
\EndIf
\State $D^i_t \gets \mathbf{d}(M^i_t)$
\EndFor
\State Construct $S_t$ from $s_t$ and $\eta_t$
\State With probability $\epsilon$, select a random action $a_{t+1}$
\State otherwise $a_{t+1} \gets \arg\max\pi(S_t)$
\State Construct $\eta_{t+1}$ with $a_{t+1}$
\State Compute $R_t(a_t)$
\State Add $(S_{t-1}, a_t, S_t, R_t)$ to replay buffer
\State $t \gets t + 1$
\EndWhile

\State Sample batch $\bf{B}$ from replay buffer
\State $\pi \gets$ \Call{DDQN}{$\bf{B}$} \Comment{Update policy using Double Q-learning}
\Until{\text{convergence}}
\end{algorithmic}
\end{algorithm}

\subsection{Inference Optimizations}
We explore two additional optimizations geared towards practical usage which triple the frame rate of our method while preserving accuracy and ease of deployment.

\begin{figure}[ht]
\begin{center}
  \includegraphics[width=0.95\linewidth]{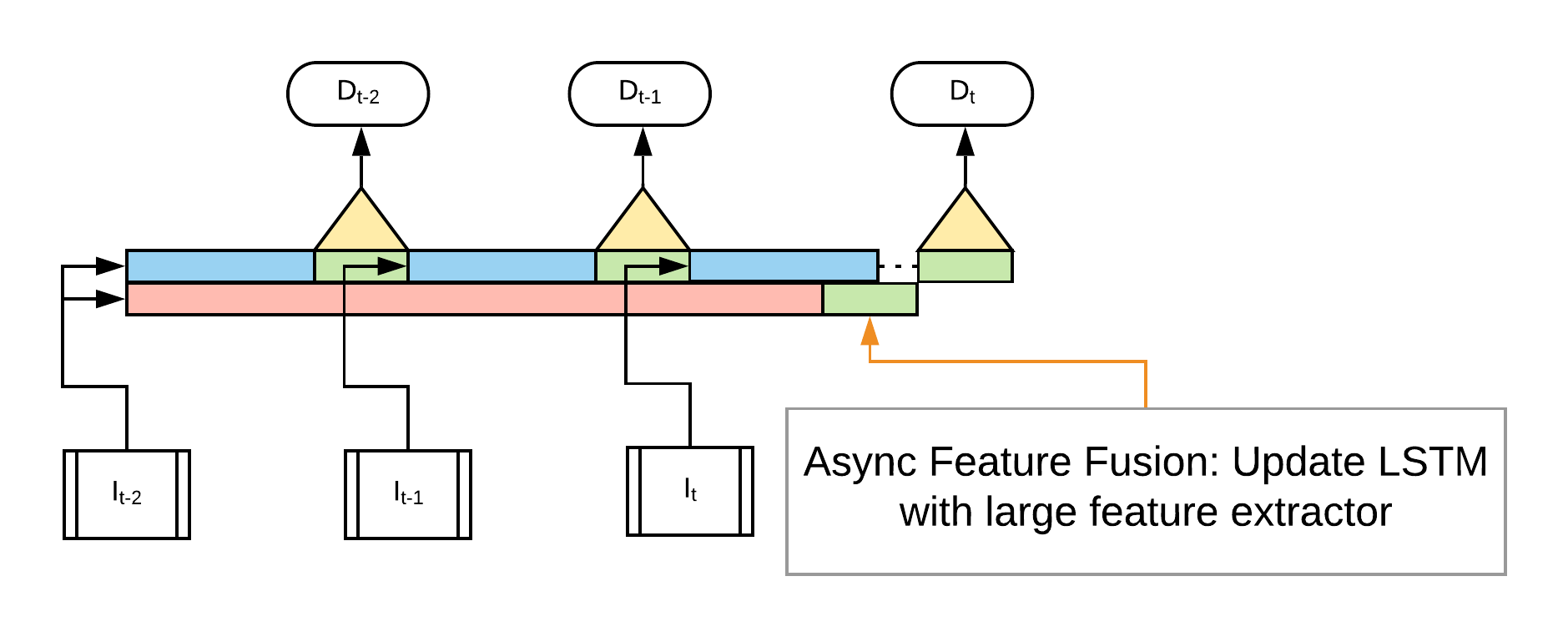}
  \vspace{-2em}
\end{center}
  \caption{Illustration of asynchronous mode.}
\label{fig:async}
\vspace{-0.5cm}
\end{figure}

\paragraph{\textbf{Asynchronous Inference}} One problem with keyframe-based detection methods is that they only consider amortized runtime. However, since these methods perform the bulk of their computation on keyframes, the latency across frames is extremely inconsistent. When worst-case runtime on a single frame is considered, these methods are no faster than single-frame methods, limiting their applicability in practical settings. Li et al. \cite{li2018low} address this problem in semantic video segmentation by running networks in parallel. Likewise, the interleaved framework naturally suggests an asynchronous inference approach which removes the gap between the amortized and worst-cases runtimes, allowing our method to run smoothly in real-time on a mobile device.

When running an interleaved model synchronously, one feature extractor is run at each timestep, so the maximum potential latency depends on $\mathbf{f_0}$. However, this process is easily parallelizable by running feature extractors in separate threads, which we term asynchronous mode. In asynchronous mode, $\mathbf{f_1}$ is run at each step and exclusively used to generate detections, while $\mathbf{f_0}$ continues to be run every $\tau$ frames and updates the memory when it completes. The lightweight feature extractor uses the most recent available memory at each step and no longer has to wait for the larger feature extractor to run. This is illustrated in Figure \ref{fig:async}.

\paragraph{\textbf{Quantization}} Another benefit of relying on multiple feature extractors instead of optical flow is that standard inference optimization methods can be applied with minimal changes. In particular, we demonstrate that our interleaved framework can be quantized using the simulated-quantization training procedure in \cite{jacob2018quantization}. The Tensorflow \cite{Abadi} quantization library is used out-of-the-box for the MobileNet and SSDLite layers. For the LSTM, fake quantization operations are inserted after all mathematical operations (addition, multiplication, sigmoid and ReLU6), following Algorithm 1 in \cite{jacob2018quantization}. Ranges after activations are fixed to $[0, 1]$ for sigmoid and $[0, 6]$ for ReLU6 to ensure that zero is exactly representable. We also ensure that the ranges of all inputs for concatenation operations are the same to remove the need for rescaling (as described in A.3. \cite{jacob2018quantization}). Our final quantized asynchronous model runs at 72.3 FPS on a Pixel 3 phone, over three times the frame rate of our unoptimized model.

\section{Experiments}
We present results on the Imagenet VID 2015 dataset \cite{Russakovsky}, which includes 30 object classes. For training, we use Imagenet VID training data and relevant classes from the Imagenet DET \cite{Russakovsky} and COCO \cite{lin2014microsoft} training sets, amounting to 3862 videos, 147K images from Imagenet DET, and 43K images from COCO. We also provide results without COCO data as \cite{zhu2018towardsmobile} does not include it, though they train on additional optical flow data. For evaluation, we use the 555 videos in the Imagenet VID validation set.

\subsection{Results on Imagenet VID}
\begin{table*}[t]
\resizebox{1.9\columnwidth}{!}{
\begin{tabular}{c | l | c | c | c | c | c }
Type & Model & Extra Data & mAP & Params (M) & MAC (M) & Runtime (FPS) \\ 
\toprule [0.2em]
\multirow{ 4}{*}{Single Frame} &  MobilenetV2-SSDLite ($\alpha=1.4$) & \multirow{ 4}{*}{COCO} & 60.5 & 6.4 & 1345 & 3.7 \\
& MobilenetV2-SSDLite ($\alpha=1.0$) &  & 58.4 & 3.5 & 710 & - \\
& MobilenetV2-SSDLite ($\alpha=0.5$) &  &  48.4 & 1.1 & 266 & - \\
& MobilenetV2-SSDLite ($\alpha=0.35$) &  & 42.0 & 0.6 & 204 & 14.4 \\
\hline
\multirow{ 4}{*}{LSTM-Based} & MobilenetV2-SSDLite + LSTM  \cite{liu2018mobile} ($\alpha=1.4$) & \multirow{ 4}{*}{COCO}  & 64.1 & 5.1 & 1182 & 4.1 \\
& MobilenetV2-SSDLite + LSTM \cite{liu2018mobile} ($\alpha=1.0$) &  & 59.1 & 2.9 & 630 & - \\
& MobilenetV2-SSDLite + LSTM \cite{liu2018mobile} ($\alpha=0.5$) &  & 50.3 & 1.1 & 255 & - \\
& MobilenetV2-SSDLite + LSTM \cite{liu2018mobile} ($\alpha=0.35$) &  & 45.1 & 0.7 & 207 & 14.6 \\
\hline
\multirow{ 3}{*}{Flow-Guided} & Zhu et al. \cite{zhu2018towardsmobile} ($\alpha=1$, $\beta=1$) & \multirow{ 3}{*}{Flying Chairs} & 61.2 & 9.0 & 205 & 12.5* \\
& Zhu et al. \cite{zhu2018towardsmobile} ($\alpha=1$, $\beta=0.5$) &  & 60.2 & 7.1 & 170 & 25.6* \\
& Zhu et al. \cite{zhu2018towardsmobile} ($\alpha=0.75$, $\beta=0.75$) &  & 56.4 & 5.3 & 115 & 26.3* \\
\hline
\multirow{ 2}{*}{Non-Interleaved (Ours)}
& Large ($\mathbf{f_0}$) only & \multirow{ 2}{*}{COCO} & 63.9 & 4.4 & 1153 & 4.2 \\
& Small ($\mathbf{f_1}$) only & & 30.6 & 1.5 & 84 & 48.8 \\
\bottomrule [0.2em]
\multirow{ 9}{*}{Memory-Guided (Ours)} 
& Interleaved & -- & 58.9 & 4.9 & 190 & 23.5 \\\cline{2-7}
& Interleaved & \multirow{ 8}{*}{COCO} & 61.4 & 4.9 & 190 & 23.5 \\
& Interleaved + Async &  & 60.4 & 4.9 & 84(190$)^{\dagger}$ & 48.8 \\\cline{2-2}\cline{4-7}
& Interleaved + Adaptive & & 61.4 & 5.0 & 168 & 26.6 \\
& Interleaved + Adaptive + Async & & 60.7 & 5.0 & 84(168$)^{\dagger}$ & 48.8 \\\cline{2-2}\cline{4-7}
& Interleaved + Quantization & & 60.3 & 4.9 & 190 & 40.8 \\
& Interleaved + Quantization + Async & & 59.3 & 4.9 & 84(190$)^{\dagger}$ & 72.3 \\\cline{2-2}\cline{4-7}
& Interleaved + Quantization + Adaptive & & 60.4 & 5.0 & 202 & 40.8 \\
& Interleaved + Quantization + Adaptive + Async & & 59.1 & 5.0 & 84(202$)^{\dagger}$ & 72.3 \\
\bottomrule [0.2em]
\end{tabular}
}\vspace{-.5em}
\centering
\caption{Results on the Imagenet VID validation set. All of our non-adaptive models use a fixed interleave policy with $\tau=9$. $\alpha$ is the feature extractor width multiplier described in \cite{Howard}, while $\beta$ is the flow network width multiplier. *Runtime results of Zhu et al. are reported with HuaWei Mate 8 phone, while the rest are reported on a Pixel 3 phone. ${}^{\dagger}$ The effective MAC for asynchronous inference  (84) for each frame includes only $\mathbf{f_1}$ plus the LSTM and SSDLite detection layers, while 190 is the amortized MAC including $\mathbf{f_0}$.}
\label{tab:main}
\end{table*}

Table \ref{tab:main} contains comparisons of our results with single-frame baselines, LSTM-based methods, and the state-of-the-art mobile video object detection method by Zhu et al. \cite{zhu2018towardsmobile}. We retrain the LSTM-based models of \cite{liu2018mobile} using publicly available code\cite{mobilelstd} on our combined dataset, including COCO data. We report accuracy in terms of mAP @0.5IOU, theoretical complexity in the form of multiply-add count (MAC), and practical runtime by deploying our model in a Pixel 3 phone using Tensorflow Lite \cite{Abadi}. Our method matches the accuracy of the most accurate flow-based mobile models with COCO data and achieves comparable accuracy even without COCO data. None of our models require any optical flow data, whereas \cite{zhu2018towardsmobile} also uses the Flying Chairs dataset with 22K examples during training. Our method also has comparable theoretical complexity, but contains far fewer parameters. 

Our adaptive model is discussed in detail in Section \hyperref[sec:4.3]{4.3}, but we include one variant ($\gamma=1$) which successfully reduces the frequency $f_0$ is run while maintaining accuracy, though it is less amenable to quantization. Our inference optimizations triple the frame rate at a small cost of accuracy, allowing our method to run much faster in practice. Though the runtimes for \cite{zhu2018towardsmobile} are measured using a different phone and not directly comparable, it is safe to say that our inference-optimized method provides unprecedented real-time runtime on mobile devices.

We also include results for only running the large and small models in our interleaved framework (i.e. $\tau=0$ and $\tau=\infty$). Our results show that both feature extractors are able to perform detection individually, but $\mathbf{f_1}$ performs extremely poorly without temporal context supplied by $\mathbf{f_0}$. This demonstrates the necessity of fusing features from both models to create a fast and accurate detector, and the effectiveness of our memory-guided framework in doing so.

\subsection{Speed/Accuracy Tradeoffs}
\begin{figure}[t]
\centering
\subfloat{\resizebox{0.8\columnwidth}{!}{
\begin{tikzpicture}
\begin{axis}[
    xlabel={Interleave Ratio ($\tau$)},
    ylabel={mAP},
    xmin=0, xmax=40,
    ymin=57, ymax=65,
    y=0.45cm,
    x=0.18cm,
    xtick={0, 5, 10, 15, 20, 25, 30, 35, 40},
    ytick={57, 59, 61, 63, 65},
    legend pos=north east,
    ymajorgrids=true,
    grid style=dashed,
]
\addplot[
    color=blue,
    mark=o,
    ]
    coordinates {
    (1,62.32)(4,61.76)(9,61.44)(19,60.27)(39,58.57)
    };
    \addlegendentry{Ours}
\addplot[
    color=red,
    mark=square,
    ]
    coordinates {
    (1, 63)(6, 62)(14, 60)(20, 58.1)
    };
    \addlegendentry{Zhu et al. \cite{zhu2018towardsmobile}}
\end{axis}
\end{tikzpicture}
}}
\centering
\vspace{-.5em}
\caption{Speed/Accuracy trade-off comparison between flow-guided and our memory-guided approach.}
\label{fig:ratio}
\end{figure}
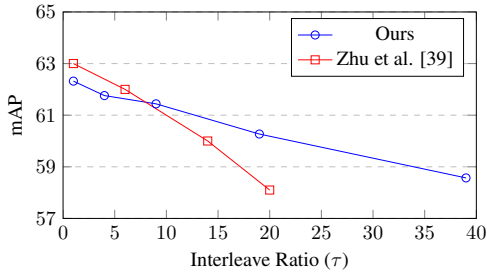

Our method provides a simple way to trade off between speed and accuracy by varying the interleave ratio $\tau$. Figure \ref{fig:ratio} plots our model's accuracy compared to the accuracy of Zhu et al.'s method using different keyframe durations. Between all interleaved models from $\tau=1$ to $\tau=39$, a drop of $3.75$ mAP is observed. Meanwhile, the flow-based method suffers a drop of at least $4.5$ mAP after only increasing the keyframe duration to $20$. Notably, several other works \cite{feichtenhofer2017detect, chen2018optimizing} have observed that it is possible to sparsely process frames on Imagenet VID using a variety of methods, suggesting that interleaving at moderate ratios is not very difficult. However, our method incurs less accuracy degradation even at extreme interleave ratios, suggesting that our memory-guided detection approach is superior at capturing the long-term temporal dynamics inherent to videos.

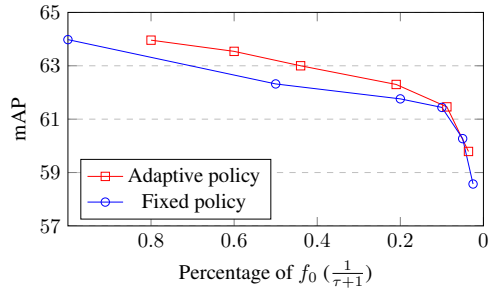
\begin{figure}[t]
\centering
\subfloat{\resizebox{0.8\columnwidth}{!}{
\begin{tikzpicture}
\begin{axis}[
    xlabel={Percentage of $f_0$ ($\frac{1}{\tau + 1}$)},
    ylabel={mAP},
    xmin=0, xmax=1,
    ymin=57, ymax=65,
    y=0.45cm,
    x=7cm,
    x dir=reverse,
    xtick={0, 0.20, 0.40, 0.60, 0.80},
    ytick={57, 59, 61, 63, 65},
    legend pos=south west,
    ymajorgrids=true,
    grid style=dashed,
]
\addplot[
    color=red,
    mark=square,
    ]
    coordinates {
    (0.8, 63.96)(0.6, 63.54)(0.44, 63.0)(0.21, 62.3)(0.088, 61.46)(0.036, 59.79)
    };
    \addlegendentry{Adaptive policy}
\addplot[
    color=blue,
    mark=o,
    ]
    coordinates {
    (1.0,63.98)(0.5,62.32)(0.2,61.76)(0.1,61.44)(0.05,60.27)(0.025,58.57)
    };
    \addlegendentry{Fixed policy}
\end{axis}
\end{tikzpicture}
}}
\centering
\vspace{-.5em}
\caption{Speed/Accuracy trade-off comparison between fixed and adaptive interleaving policies.}
\label{fig:rlgraph}
\end{figure}

\subsection{Reinforcement Learning}
\label{sec:4.3}
\begin{figure*}[!htb]
\centering
\subfloat[Q-values]{ \includegraphics[width=0.7\columnwidth]{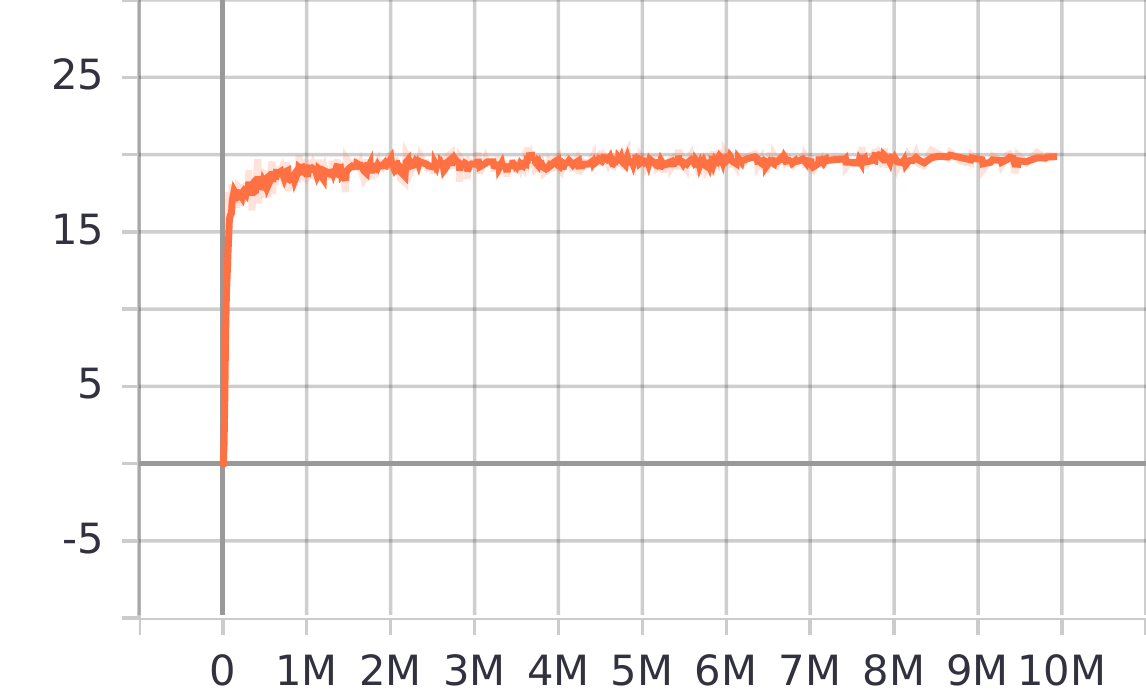}}
\subfloat[mAP]{\includegraphics[width=0.7\columnwidth]{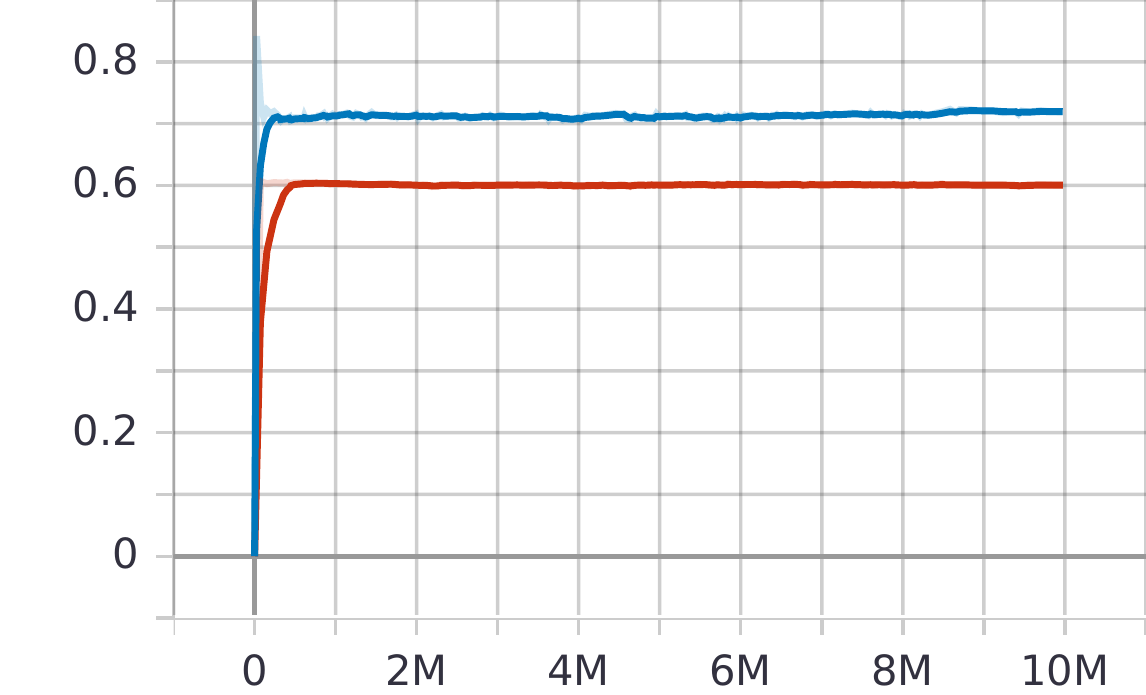}}
\subfloat[Percentage of $f_0$]{\includegraphics[width=0.7\columnwidth]{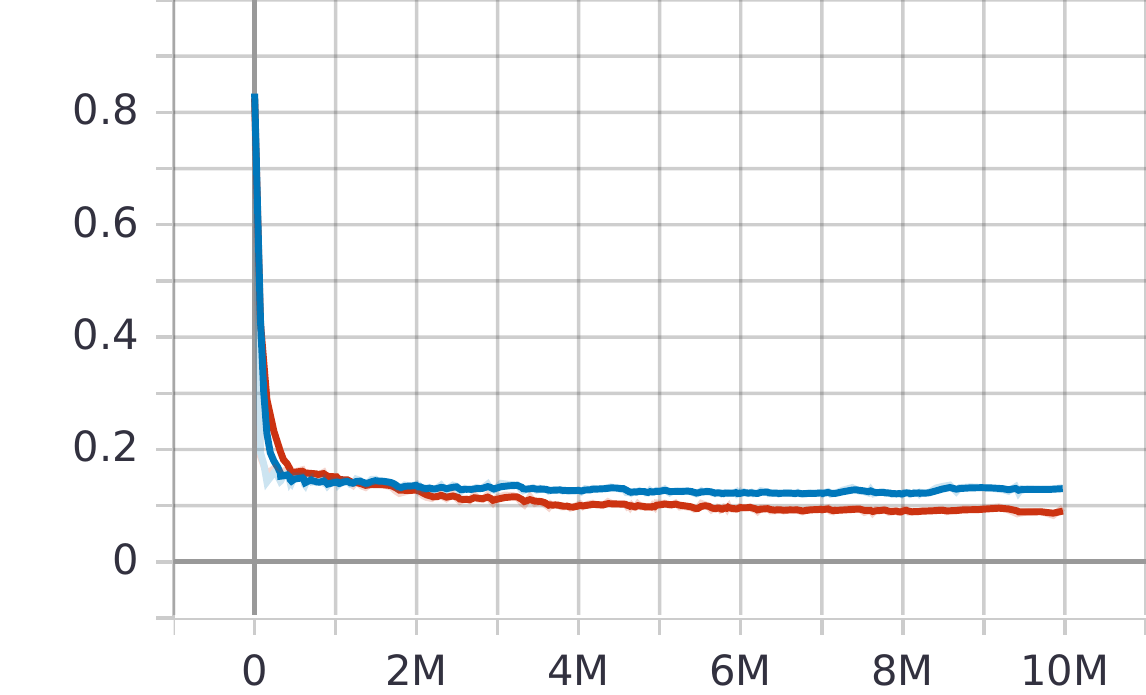}}
\caption{Q-values, mAP and percentage of $f_0$ run during RL training. The blue curves correspond to training and the red curves evaluation. The x-axis is the number of training iterations.}
\label{fig:RL_training}
\end{figure*}

\begin{figure*}[!htb]
\centering
\resizebox{0.9\linewidth}{!}{
\subfloat{\includegraphics[width=.16\textwidth,height=.12\textwidth]{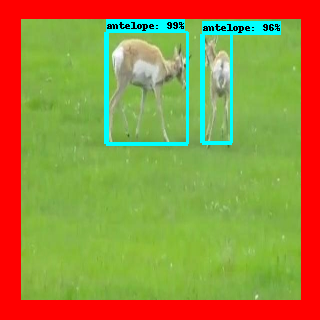}}
\subfloat{\includegraphics[width=.16\textwidth,height=.12\textwidth]{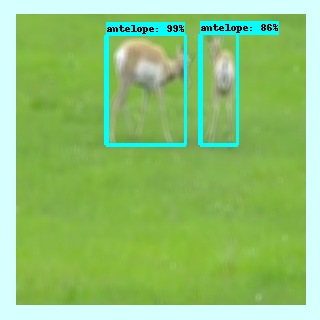}}
\subfloat{\includegraphics[width=.16\textwidth,height=.12\textwidth]{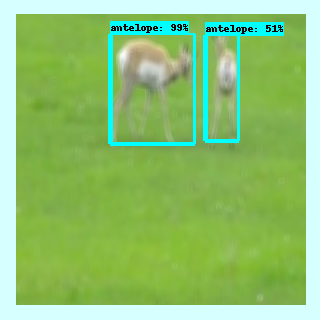}}
\subfloat{\includegraphics[width=.16\textwidth,height=.12\textwidth]{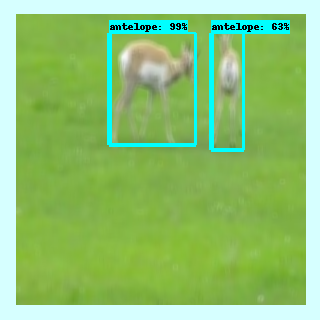}}
\subfloat{\includegraphics[width=.16\textwidth,height=.12\textwidth]{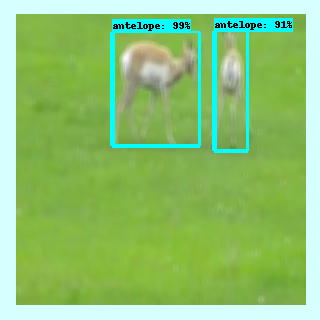}}
\subfloat{\includegraphics[width=.16\textwidth,height=.12\textwidth]{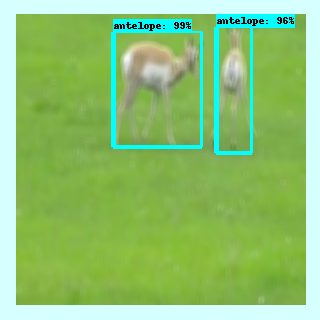}}}\\\vspace{-.5em}
\resizebox{0.9\linewidth}{!}{
\subfloat{\includegraphics[width=.16\textwidth,height=.12\textwidth]{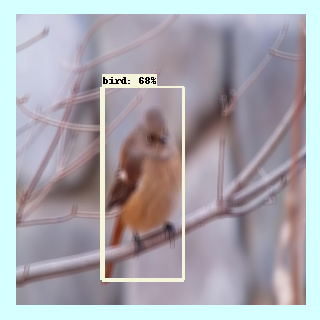}}
\subfloat{\includegraphics[width=.16\textwidth,height=.12\textwidth]{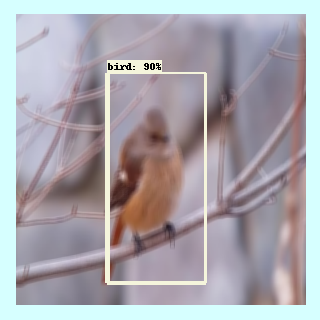}}
\subfloat{\includegraphics[width=.16\textwidth,height=.12\textwidth]{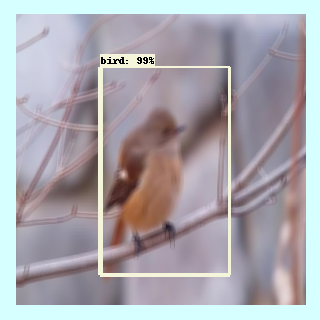}}
\subfloat{\includegraphics[width=.16\textwidth,height=.12\textwidth]{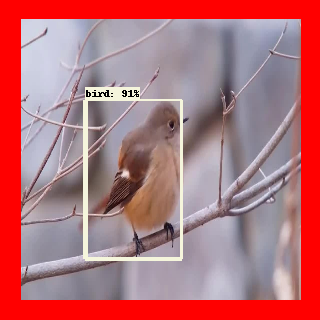}}
\subfloat{\includegraphics[width=.16\textwidth,height=.12\textwidth]{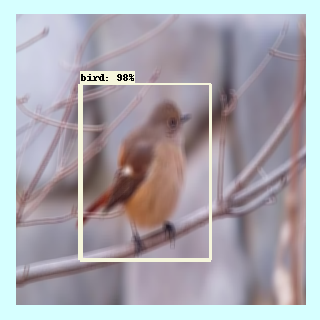}}
\subfloat{\includegraphics[width=.16\textwidth,height=.12\textwidth]{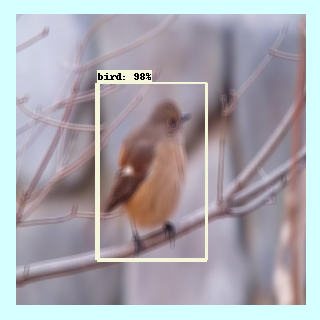}}}\\\vspace{-.5em}
\resizebox{0.9\linewidth}{!}{
\subfloat{\includegraphics[width=.16\textwidth,height=.12\textwidth]{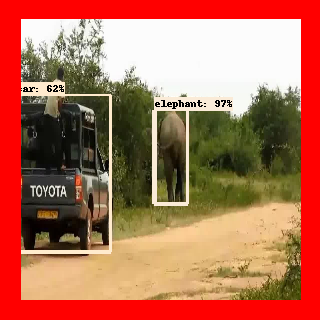}}
\subfloat{\includegraphics[width=.16\textwidth,height=.12\textwidth]{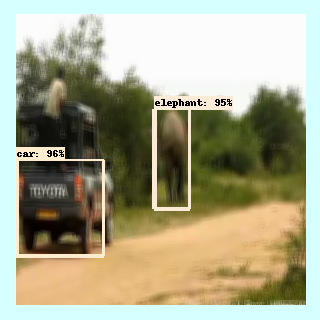}}
\subfloat{\includegraphics[width=.16\textwidth,height=.12\textwidth]{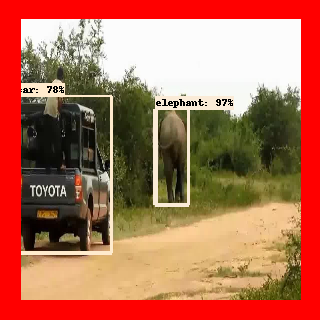}}
\subfloat{\includegraphics[width=.16\textwidth,height=.12\textwidth]{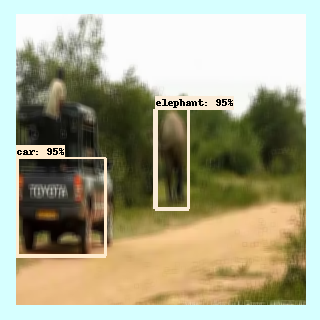}}
\subfloat{\includegraphics[width=.16\textwidth,height=.12\textwidth]{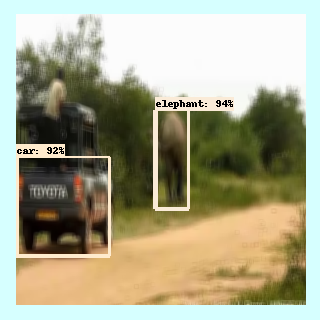}}
\subfloat{\includegraphics[width=.16\textwidth,height=.12\textwidth]{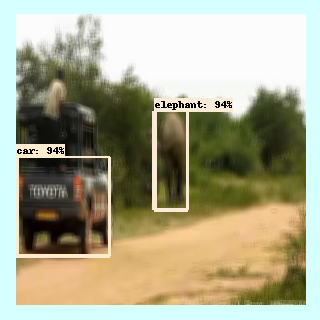}}}\\\vspace{-.5em}
\resizebox{0.9\linewidth}{!}{
\subfloat{\includegraphics[width=.16\textwidth,height=.12\textwidth]{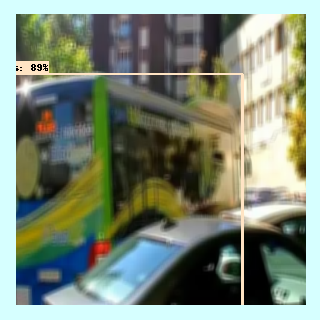}}
\subfloat{\includegraphics[width=.16\textwidth,height=.12\textwidth]{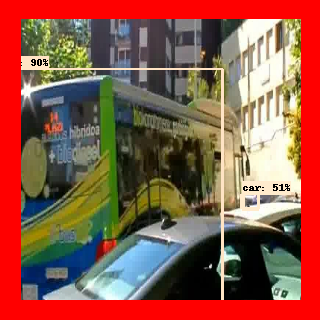}}
\subfloat{\includegraphics[width=.16\textwidth,height=.12\textwidth]{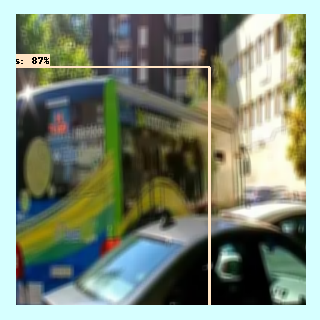}}
\subfloat{\includegraphics[width=.16\textwidth,height=.12\textwidth]{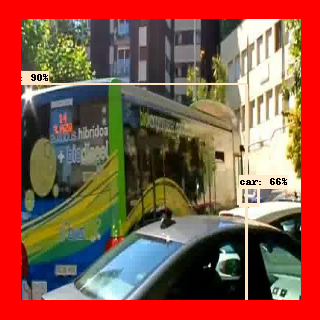}}
\subfloat{\includegraphics[width=.16\textwidth,height=.12\textwidth]{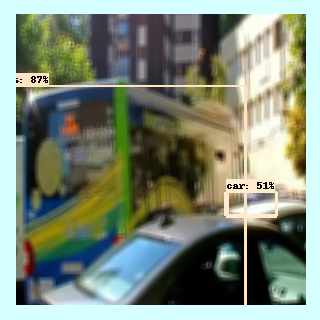}}
\subfloat{\includegraphics[width=.16\textwidth,height=.12\textwidth]{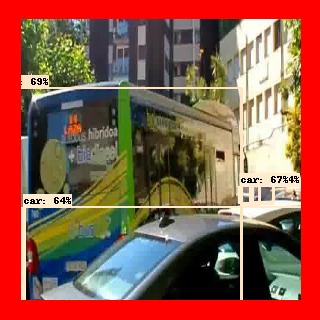}}}
\\\vspace{-0.5em}
\resizebox{0.9\linewidth}{!}{
\subfloat{\includegraphics[width=.16\textwidth,height=.12\textwidth]{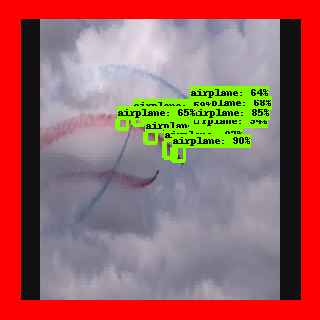}}
\subfloat{\includegraphics[width=.16\textwidth,height=.12\textwidth]{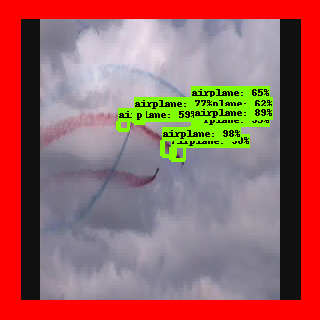}}
\subfloat{\includegraphics[width=.16\textwidth,height=.12\textwidth]{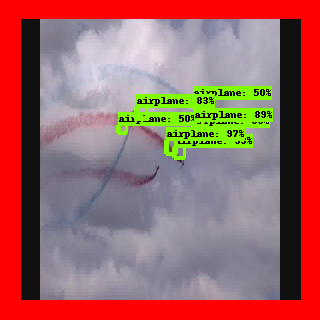}}
\subfloat{\includegraphics[width=.16\textwidth,height=.12\textwidth]{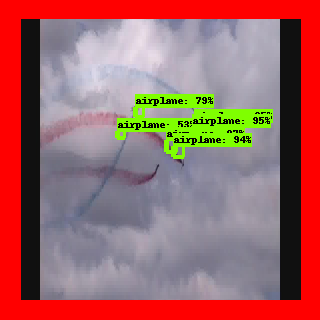}}
\subfloat{\includegraphics[width=.16\textwidth,height=.12\textwidth]{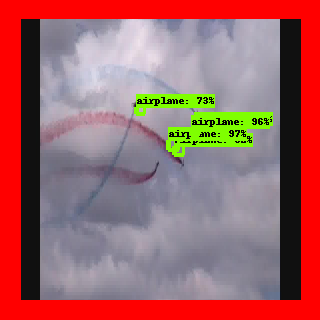}}
\subfloat{\includegraphics[width=.16\textwidth,height=.12\textwidth]{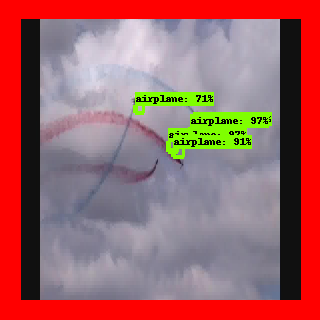}}
}
\caption{Visualization of a learned adaptive policy for feature extractor selection on Imagenet VID validation (best viewed in color). Frames where the policy runs the large model are highlighted in red. Clips are ordered by how often the large model triggers. The scene complexity increases correspondingly, which shows that the policy allocates computation intelligently.}
\label{fig:visualization}
\end{figure*}

We present results for a variety of learned interleaving policies in Figure \ref{fig:rlgraph}. We are able to vary how often $\mathbf{f_0}$ is run by adjusting the speed reward $\gamma$, with $\gamma \in \{1.5, 1.0, 0.4, 0.3, 0.2, 0.1\}$. The adaptive policy is superior to the fixed policy at all percentages, especially when more large models can be run. This improvement comes at a negligible cost of $89.6$K extra parameters and $1.76$M multiply-adds from the policy network. Figure \ref{fig:RL_training} shows the change in predicted Q-values, mAP and percentage of $f_0$ run during training. Notably, the mAP reaches a constant value, while Q-values steadily increase and interleave ratio (percentage) decreases. This observation suggests the policy network gradually learns how to use the small model without hurting the overall accuracy. Figure \ref{fig:visualization} visualizes the adaptive policy on the Imagenet VID validation set, ordered by how often the large model is run. The learned policy tends to expend more computation when objects are difficult to detect while running the small network frequently on easy scenes, demonstrating the merits of this approach over a fixed policy.

\section{Conclusion}
We propose a method for video object detection by interleaving multiple feature extractors and aggregating their results in memory. By interleaving extremely lightweight and conventional feature extractors, we construct a model optimized for computationally constrained environments. The presence of the memory module allows the lightweight feature extractor, which performs poorly by itself, to be run frequently with minimal accuracy loss. We also propose a method for learning the interleaving policy using reinforcement learning. We demonstrate that our method is competitive with the state-of-the-art for mobile video object detection while enjoying a substantial speed advantage and removing the dependency on optical flow, making it effective and straightforward to deploy in a mobile setting.

\clearpage

{\small
\bibliographystyle{ieee}
\bibliography{paper}
}
\end{document}